\documentclass{bmvc2k}
\usepackage{booktabs}
\usepackage{multirow}
\usepackage{pifont}

\title{Detecting Text Manipulation in Images using Vision Language Models}

\addauthor{Vidit Vidit}{vidit.vidit@idiap.ch}{1}
\addauthor{Pavel Korshunov}{pavel.korshunov@idiap.ch}{1}
\addauthor{Amir Mohammadi}{amir.mohammadi@idiap.ch}{1}
\addauthor{Christophe Ecabert}{christophe.ecabert@idiap.ch}{1}
\addauthor{Ketan Kotwal}{ketan.kotwal@idiap.ch}{1}
\addauthor{S\'{e}bastien Marcel}{sebastien.marcel@idiap.ch}{1}
\addinstitution{
IDIAP Research Institute\\ Martigny \\ Switzerland
}

\runninghead{Vidit et. al.}{Detecting Text Manipulation in Images using VLM}

\usepackage[capitalize]{cleveref}
\crefname{section}{Sec.}{Secs.}
\Crefname{section}{Section}{Sections}
\Crefname{table}{Table}{Tables}
\crefname{table}{Tab.}{Tabs.}
\usepackage[skip=2pt]{caption}
\usepackage[most]{tcolorbox}

\begin{document}

\maketitle

\begin{abstract}
    Recent works have shown the effectiveness of Large Vision Language Models (VLMs or LVLMs) in image manipulation detection. However, text manipulation detection is largely missing in these studies. We bridge this knowledge gap by analyzing closed- and open-source VLMs on different text manipulation datasets. Our results suggest that open-source models are getting closer, but still behind closed-source ones like GPT-4o. Additionally, we benchmark image manipulation detection-specific VLMs for text manipulation detection and show that they suffer from the generalization problem. We benchmark VLMs for manipulations done on in-the-wild scene texts and on fantasy ID cards, where the latter mimic a challenging real-world misuse. Paper Page: \url{https://www.idiap.ch/paper/textvlmdet/}
\end{abstract}
\section{Introduction}
Advancements in image generation has made it easier to edit and generate realistic looking images. However, this capability comes with a major downside, the same tools can be exploited for malicious purposes, such as  generating fake and manipulated images. Such images are increasingly being used to spread misinformation~\cite{dufour2024ammeba} or create fraudulent identities that bypass online Know Your Customer (KYC) checks\footnote{https://securityexpress.info/fake-passport-generated-by-chatgpt-bypasses-security/}. Typically, these manipulations involve addition/removal of objects in an image through a generative method and can be followed by post-processing to seamlessly blend the altered regions. With rapid progress in generation quality, these manipulated images have become more difficult to visually identify, especially, when small but semantically critical regions, such as text, are modified. Detecting such subtle changes is challenging and current image forgery detection methods often overlook the manipulated text regions. 
One of the goals of this work is to bridge this gap.
\begin{figure}[ht]
    \centering
    \includegraphics[width=\textwidth]{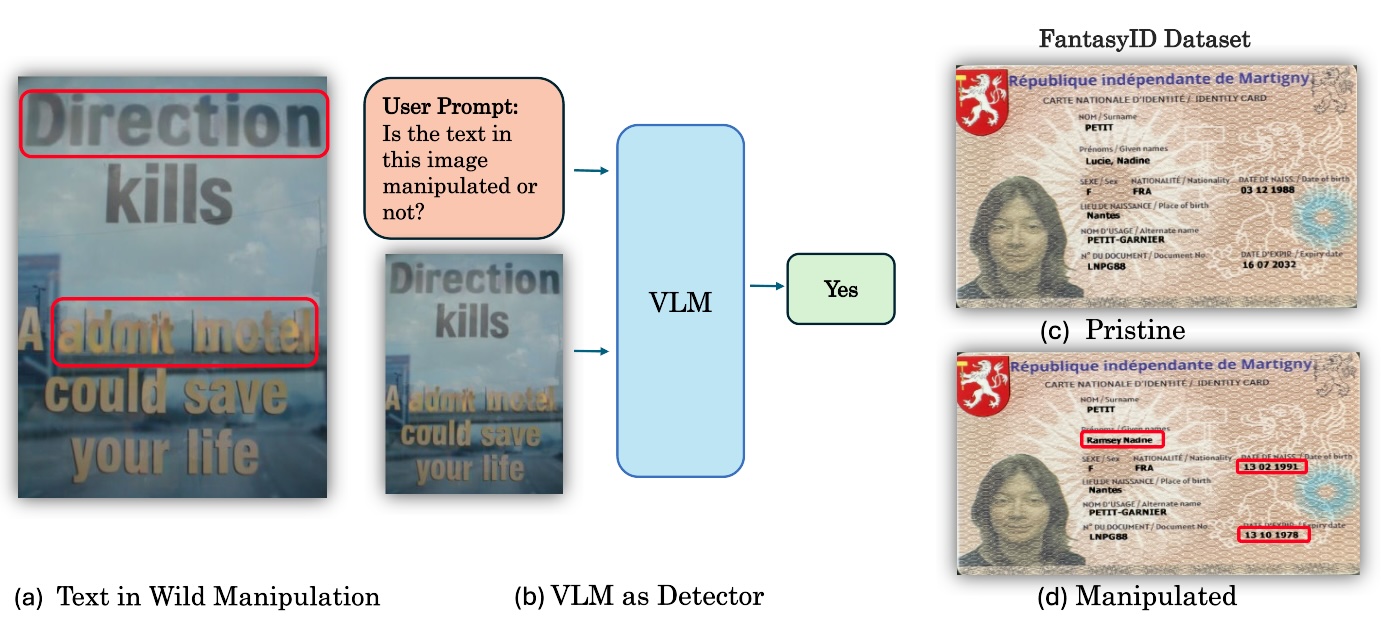}
    \caption{\textbf{VLM for Text Manipulation Detection:} (a) Example from OSTF~\cite{qu2025revisiting} dataset of in-the-wild text manipulation (in \colorbox{red!60}{red}). (b) With the help of a user prompt, we ask a pretrained VLM whether the accompanying image contains a text manipulation. The output of VLM is then used as a label for binary classification. (c,d) Example from FantasyID~\cite{korshunov2025fantasyid} dataset which simulate the real-world scenario of text manipulation in ID documents. (d) The altered text is shown in \colorbox{red!60}{red}. Best viewed digitally.}
    \label{fig:teaser}
\end{figure}

Recent works have shown promising results in the detection of generated and manipulated images~\cite{guillaro2023trufor,li2019localization,salloum2018image,li2018fast,kwon2021cat,chen2021image,ying2023learning,yu2023cross}. Typically, trained models can provide an image-level label and/or localize tampered regions within the image. With the growing adoption of large vision language models (VLMs), they started being used to detect generated or tampered images~\cite{xu2024fakeshield,huang2024sida,komaty2025chatgpt}. One useful aspect of such models is their zero-shot reasoning ability, as well as their capacity to generate textual explanations for their prediction. VLM-based methods have shown significant improvement over smaller convolutional or transformer-based manipulation detection models for face presentation attack detection~\cite{Ozgur_2025_WACV}.

As VLMs are being employed for a variety of tasks, it becomes pertinent to understand where they stand on text manipulation detection as well. In this work, we would like to answer few questions like \emph{how big is the gap between open and closed source models?} and \emph{how well do the finetuned VLMs generalize to text manipulation detections?}. We also illustrate how the prompt design and input image resolution affect the final result. These takeaways can help in designing a better text manipulation detectors.

Although there are existing datasets with text manipulation of images taken in the wild (OSTF~\cite{qu2025revisiting}), the main focus of this work is on detecting the real-world text forgery in ID documents, since it is a critical security risk in the prevalence of online-based enrollment and KYC checks. For this purpose, evaluate how practical are current VLMs for detecting tampered text in ID documents, using FantasyID~\cite{korshunov2025fantasyid}. This data set consists of the manipulation of text fields on \emph{fantasy} id cards, which simulates an \emph{injection-attack} by a fraudster.

To summarize our contributions, (a) we study open and closed source VLMs for text manipulation detection task in zero-shot setting, (b) we evaluated existing image manipulation specific finetuned VLMs on their generalization capability to text manipulation detection, and (c) to promote research in real-world forgery detection, we benchmark different VLMs on FantasyID~\cite{korshunov2025fantasyid} and OSTF~\cite{qu2025revisiting}. We will make our code public upon acceptance.

\section{Related Work}

\paragraph{Image Manipulation Datasets.}
Several training and benchmark datasets have been proposed for image manipulation detection. Common datasets include Photoshop-edited and deepfake collections such as CASIAv2~\cite{dong2013casia}, FantasticReality~\cite{kniaz2019point}, Columbia~\cite{ng2009columbia}, COVERAGE~\cite{wen2016coverage}, IMD2020~\cite{novozamsky2020imd2020}, and FaceApp~\cite{dang2020detection}. These datasets offer limited manipulation diversity and are primarily used for evaluation rather than large-scale training. FakeShield~\cite{xu2024fakeshield} and SIDA~\cite{huang2024sida} introduce large-scale AI-generated image datasets (based on GANs or diffusion models) for training. However, they lack sufficient evaluation of text-based manipulations. Open-Set Text Forensics (OSTF)~\cite{qu2025revisiting} addresses this by introducing a dataset with text edits generated using both CNN and diffusion models. We also study ID card datasets to better understand detection performance on the real-world document tampering. PAD~\cite{tapia2024first} propose ID card datasets
for presentation attack detection scenarios. In contrast, we target digital manipulation reflecting an injection attack scenario. While PAD~\cite{tapia2024first} involves manipulation of original government IDs, which is legally prohibited, FantasyID~\cite{korshunov2025fantasyid} avoids legal issues by creating fantasy templates and using images with appropriate licenses. In this work, we evaluate detection methods on both the OSTF dataset and FantasyID. 

\begin{figure}[t]
    \centering
    \includegraphics[width=\linewidth]{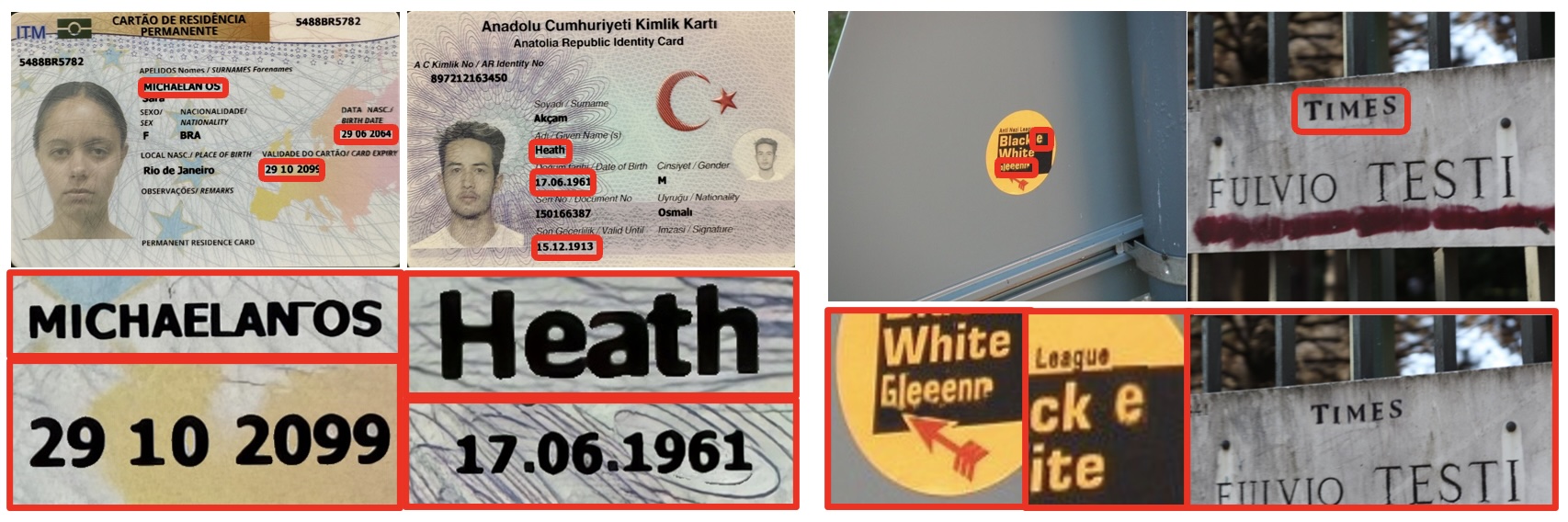}
    \caption{\textbf{FantasyID~\cite{korshunov2025fantasyid} and OSTF~\cite{qu2025revisiting}} Two different templates from our proposed, FantasyID (left) and OSTF (right) dataset. The \colorbox{red!60}{red} box show the manipulated regions and their corresponding zoomed-in image. (Left) Altered text mostly show jagged edges due to inconsistent blending with the background. (Right) Zoomed-in view of the tampered region where some of the manipulation artifacts are visible. Such regions occupy small area of the full image making it a challenging detection task.  }
    \label{fig:fantasy_example}
\end{figure}
\vspace{-1.5em}
\paragraph{Image Manipulation Detection.}
Various works~\cite{guillaro2023trufor,li2019localization,salloum2018image,li2018fast,kwon2021cat,chen2021image,ying2023learning,yu2023cross} have proposed detection methods targeting different types of forgeries, such as \emph{inpainting}, \emph{copy-move}, and \emph{splicing}. Detection tasks can involve image-level classification or localization of tampered regions. Typically, training datasets contain pristine and manipulated images, with or without annotations indicating the manipulated areas. These manipulations are introduced using CNN-based generative methods or diffusion-based approaches. TruFor~\cite{guillaro2023trufor} learns camera-specific noise patterns and identifies tampered regions by detecting inconsistencies in these patterns. HiFi-Net~\cite{guo2023hierarchical} detects image alterations using multi-branch feature extraction. MVSS-Net~\cite{dong2022mvss} utilizes multi-scale, multi-branch features to better capture image noise, while IML-ViT~\cite{ma2023iml} employs a transformer-based architecture with edge-guided loss to enhance focus on tampering artifacts. EditGuard~\cite{zhang2024editguard} embeds watermarks into images and later verifies authenticity based on their presence. These methods are trained on specific types of tampering and aim to generalize to unseen manipulation techniques. However, as generative models improve, generalizing to novel manipulations becomes increasingly challenging. Text tampering detection methods~\cite{wang2022detecting,qu2023towards,qu2025revisiting} highlight the importance of benchmarks tailored to text manipulation, as such edits often affect small image regions. Nevertheless, similar to prior approaches, these methods also struggle to generalize across new editing techniques.
\vspace{-1.5em}
\paragraph{Vision-Language Models for Detection.}
The use of large Vision-Language Models (VLMs) has grown rapidly, largely due to their zero-shot capabilities. Trained on large-scale internet datasets, these models can quickly adapt to downstream applications. While proprietary models such as GPT-4o\footnote{\url{https://platform.openai.com/docs/models/gpt-4o} } have demonstrated strong performance on a range of benchmarks~\cite{chiang2024chatbot}, open-source alternatives like Qwen-VL~\cite{Qwen2.5-VL} and Llama-Vision~\cite{grattafiori2024llama}
are closing the gap. These transformer-based models tokenize images into patch sequences and project them into a shared embedding space with text features, enabling rich cross-modal representations. Recently, VLMs have been applied to detect and localize image manipulations with FakeShield~\cite{xu2024fakeshield} and SIDA~\cite{huang2024sida}, fine-tuned LLaVA~\cite{liu2023visual} models, 
not only achieving state-of-the-art performance but also generating textual justifications for their predictions, enhancing interpretability. Here, we study these VLMs for text manipulation detection which is largely missing in their evaluations.

\section{Experimental Setup}

We evaluate GPT-4o which is a closed source VLM, Qwen-VL~\cite{Qwen2.5-VL} and Llama-Vision-3.2~\cite{grattafiori2024llama} which are generic open source VLM, through zero-shot prompting task. As shown in~\cref{fig:teaser}, we prompt query image with a detailed prompt asking VLM to generate a final label. We further study the finetuned VLMs, FakeShield~\cite{xu2024fakeshield} and SIDA~\cite{huang2024sida},  on their generalization to text manipulation detection. We use their default prompts to get the prediction of the query image. Additionally, we compare with a non-VLM baseline TruFor~\cite{guillaro2023trufor}, because it has a good generalization capability to unseen tampering methods. 

\subsection{Datasets}
Most VLM-based manipulation detection methods lack a study on text specific manipulations. This task is challenging because usually text regions occupy smaller area in the image (\cref{fig:fantasy_example}). Therefore, we present their zero-shot evaluation on the following datasets:
\vspace{-1.5em}
\paragraph{Open-Set Text Forensics (OSTF)~\cite{qu2025revisiting}} dataset consists of images with in-the-wild text manipulation, which are sourced from $4$ different scene text datasets, and manipulation is done using $8$ different methods comprising text rendering, a convolution, and diffusion model-based editing. We use their test split with $2238$ images (1267 pristine, 971 manipulated) for our evaluation. 
\vspace{-1.5em}
\paragraph{FantasyID~\cite{korshunov2025fantasyid}}Text manipulation in sensitive documents poses severe security risks for KYC applications. 
This dataset has $362$ ID cards which mimic real-world IDs. IDs are printed on PVC cards and captured using $3$ camera devices. Unlike OSTF~\cite{qu2025revisiting}, the text-manipulated images by fine-tuning TextDiffuser-2~\cite{chen2024textdiffuser} on pristine cards. This creates difficult to detect manipulation as the artifacts (\cref{fig:fantasy_example}) are subtle.  For our evaluations, we use the $1572$ images ($786$ pristine, $786$ manipulated). 

\subsection{Baselines}\label{subsec:baselines} 
\paragraph{TruFor~\cite{guillaro2023trufor}}uses a multi-branch Transformer encoder architecture to combine features from RGB images
and Noiseprint++ images to predict an anomaly localization map, a confidence map, and a final score. Noiseprint++~\cite{guillaro2023trufor} is a fully convolutional network trained to extract subtle noise in pristine images caused by imperfections in camera hardware or in-camera processing steps. When an image is manipulated, it creates a different noise pattern in the manipulated region, which can be treated as an anomaly. This is used to predict a final score for the image. Trufor can accept images of arbitary sizes but we consider two variants (a) TruFor-\emph{low}  longest side is resized to $512$ (b) TruFor-\emph{high}, the longest side is resized to $768$ only for the images that cannot be processed in the original image dimension.
\vspace{-1.5em}
\paragraph{SIDA~\cite{huang2024sida}} is based on LISA~\cite{lai2024lisa} model, which the authors fine-tuned on a large dataset of manipulated images collected from social media. The model generates an explanation of manipulation along with its mask. We use the label provided by the model for evaluation. We use the default prompt with the query image. For input, the images are split into $4$ non-overlapping patches of size $336^2$ and then processed by vision encoder.
\begin{figure}[t]
    \centering
    \includegraphics[width=\linewidth]{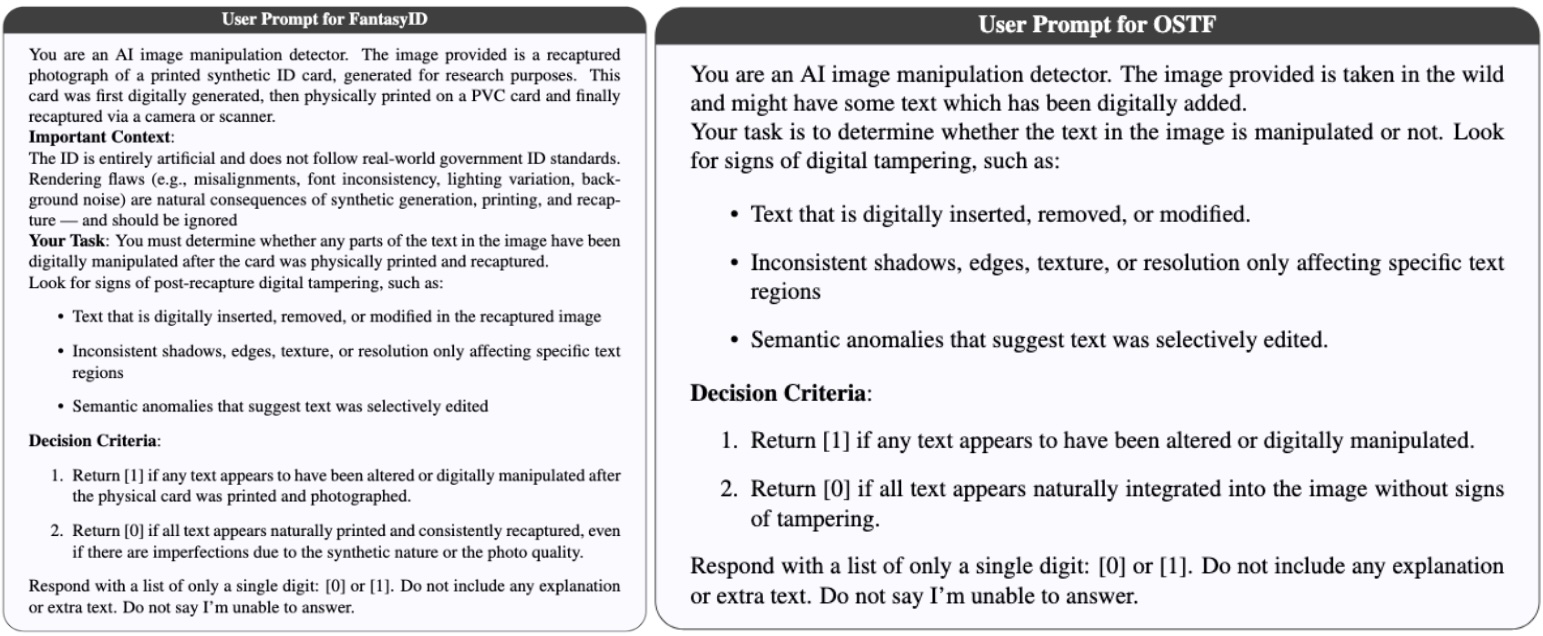}



    


    \caption{\textbf{VLM Prompt for FantasyID~\cite{korshunov2025fantasyid} and OSTF~\cite{qu2025revisiting}} We provide the necessary context of the problem along with intended output format.}
    \label{fig:ostf-prompt}
\end{figure}
\vspace{-1.5em}
\paragraph{FakeShield~\cite{xu2024fakeshield}} proposes a fine-tuned LLaVA-v$1.5$-$13$B~\cite{liu2023visual} model  for manipulation detection and localization, which is tuned on multimodal dataset of  manipulated images to generate description of manipulated region along with its localization mask. The training data consists of edited images based on deepfakes, photoshop, and diffusion models. In our evaluations, we use the default prompt and consider an image manipulated when the model generates a mask. The input image size is the same as in SIDA.
\vspace{-1.5em}
\paragraph{Llama-Vision} We benchmark the open-source Llama-$3.2$-$90$B model with vision capability in the zero-shot setting. It is trained on general image reasoning tasks with webscale data, and we prompt this model to behave as a text manipulation detector with temperature set to $0$ for \emph{deterministic} answer. The input image are split into up to $4$ square patches of size $560$.
\vspace{-2em}
\paragraph{Qwen-VL~\cite{Qwen2.5-VL}} is another powerful open-source general VLM. Similarly to Llama, we prompt this model (Qwen$2.5$-VL-$72$B version) to behave as a text manipulation detector in the zero-shot setting. 
While we keep the default image resolution, we set the model temperature to $0$ to provide a \emph{deterministic} answer. Qwen model can take images of arbitrary sizes and process them as a sequence of $28^2$ non-overlapping patches.
\vspace{-1.5em}
\paragraph{GPT-$4$o} We evaluate the closed-source GPT-$4$o  model (\texttt{gpt-4o-2024-11-20} version), 
for the manipulation detection task. OpenAI's API allows a query with images in two resolutions: \emph{low} and \emph{high}. The former process image resolution with the longest side of $512$, while the latter can handle $768 \text{(short-side)} \times 2000 \text{(long-side)}$. The images are dynamically resized or padded to fit these dimensions while maintaining the aspect ratio. The model temperature is set to $0$ to provide a \emph{deterministic} answer.

\subsection{Metrics and compute details}
We measure binary classification performance using $F_1$ scores and Avg $F_1$. The confidence threshold of $0.5$ is set for TruFor baselines. We do not measure localization accuracy as it is non-trivial to convert the output from GPT-4o, Qwen and Llama into a mask. This mask generation is beyond the scope of the work. 
For open-source models, we use a single H100 GPU, and GPT-4o was accessed through the OpenAI API, whose total experiment
cost was around \$200.

\subsection{Prompt Design}
For GPT-4o, Llama-Vision, and Qwen-VL, we take a zero-shot approach in which, given a prompt and a query image, VLM provides a final prediction for the image. This prediction is then used as a label in binary classification. We need to design an appropriate prompt that describes the underlying task and the required output format. The prompts used for the experiment are shown in ~\cref{fig:ostf-prompt}. FantasyID prompts are more detailed than OSTF as they are \emph{fantasy} IDs and can be considered as fake, if not explicitly prompted to ignore that fact. The prompts for FakeShield~\cite{xu2024fakeshield} and SIDA~\cite{huang2024sida} are the same as proposed by the authors. Since these are specialized for manipulation detection task, they do not need different prompts.

\section{Results}
\begin{table}[t]
    \centering
    \begin{tabular}{l|ccc|ccc}
    \toprule
    \multirow{2}{*}{Model} & \multicolumn{3}{c|}{OSTF} & \multicolumn{3}{c}{FantasyID} \\
    \cmidrule(lr){2-4} \cmidrule(lr){5-7}
     &  $F_1(P)$ & $F_1(M)$ & Avg $F_1$ & $F_1(P)$ & $F_1(M)$ & Avg $F_1$ \\
    \midrule
        TruFor-low~\cite{guillaro2023trufor} & 0.72  &     0.17  & 0.45 &    0.64  &     0.20  &     0.42   \\
        TruFor-high~\cite{guillaro2023trufor} & 0.74  &    0.56  &    0.65 & 0.70  &    0.72  &    0.71 \\
        FakeShield~\cite{xu2024fakeshield}  & 0.51  &    0.51  &     0.51 & 0.51  &    0.51  &   0.51  \\ 
        SIDA~\cite{huang2024sida} & 0.70   &   0.24  &    0.47 & 0.67   &   0.01  &    0.34 \\ 
        Llama-3.2-90b-Vision  & 0.75  &    0.52   &   0.64 & 0.65   &   0.27 &    0.46  \\ 
        Qwen-2.5-VL-72b~\cite{Qwen2.5-VL}  & 0.85  &    0.74   &   0.79 & 0.72  &    0.40  &    0.56  \\ 
        GPT-4o-low  & 0.87  &    0.82  &    0.84& 0.74 &     0.50 &     0.62 \\ 
        GPT-4o-high & 0.86 &  0.85 & \textbf{0.86} & 0.84 &	0.86 &	\textbf{0.85} \\
    \bottomrule
    \end{tabular}
    \vspace{0.5em}
    \caption{\textbf{Evaluation on OSTF~\cite{qu2025revisiting} and FantasyID~\cite{korshunov2025fantasyid}:} We compare the performance of different VLMs and a non-VLM baseline TruFor using Avg.$F_1$. 
    Note: \textbf{$F_1(P)$}: $F_1$ for Pristine class, \textbf{$F_1(M)$}: $F_1$ for Manipulated class }
    \label{tab:res_ostf}
\end{table}

\paragraph{Text Manipulation Detection.} We evaluate our baselines with the designed prompts and ~\cref{tab:res_ostf} report the performance on the two datasets showing GPT-4o to be the best performing model  overall.
\begin{itemize}
    \item \emph{closed vs open source VLM:} Both Qwen-VL and Llama-3.2-Vision models underperform compared to GPT-4o.  Avg $F_1$ for GPT-4o is higher  than  Qwen-VL by $29$\% on FantasyID and $7$\% on OSTF dataset~\cref{tab:res_ostf}. Despite having access to the original image resolution,  Qwen-VL  still underperforms even compared to GPT-4o-low version, which often downsizes input images. Among open-source models, Qwen-VL performs consistently better than Llama-3.2-Vision, similar to other reported benchmarks~\cite{chiang2024chatbot}. Clearly, the open source models are lagging behind ChatGPT but Qwen-VL can be a competitive choice. 
    \item \emph{general vs specialized VLM:} FakeShield~\cite{xu2024fakeshield} and SIDA~\cite{huang2024sida} underperform compared to all the general VLMs: Llama, Qwen and GPT-4o. The reason could be that specialized VLMs are finetuned versions of a smaller LLaVa-13B~\cite{liu2023visual} model. FakeShield and SIDA are limited by both the reasoning capacity of the smaller LLM and the vision encoder, which can only handle smaller image resolution. However, they had competitive results on their own scene/object manipulation benchmarks~\cite{huang2024sida,xu2024fakeshield}, which also suggests that they suffer from the generalization problem. In their training and evaluation datasets, they rarely see text manipulation, making them biased towards artifacts of the scene manipulations.

    \item \emph{VLM vs TruFor:} TruFor~\cite{guillaro2023trufor} is a non-VLM baseline with a good out of distribution generalization. Here we see that compared to GPT-4o, it underperforms on both low and high versions. However, it is better than FakeShield~\cite{xu2024fakeshield} and SIDA~\cite{huang2024sida} on OSTF but similar on FantasyID. It shows that TruFor, while being a smaller model, can be competitive when compared to specialized VLMs. It is worse than Qwen-VL and Llama-Vision on OSTF but better on FantasyID when using higher image resolution. We would like to emphasize that TruFor is trained to detect manipulation, whereas Qwen-VL and Llama-Vision are zero-shot prompted. While being competitive with `high' resolution, TruFor suffers under `low' image resolution. This results in a random accuracy for FantasyID with tiny regions of manipulated text, demonstrating the challenges of manipulation detection. 
     
\end{itemize}

In summary, closed VLM GPT-4o shows significantly higher accuracy on both datasets compared to both open and specialized VLMs. Qwen-VL remains a competitive open source option, but text manipulation detection remains a hard problem for specialized VLMs and TruFor to generalize. However, TruFor performs better than specialized VLMs on both datasets. For fair comparison, we do not include OSTF~\cite{qu2025revisiting} as it is a purely localization baseline.
\begin{figure}[t]
    \centering
    \includegraphics[width=\linewidth]{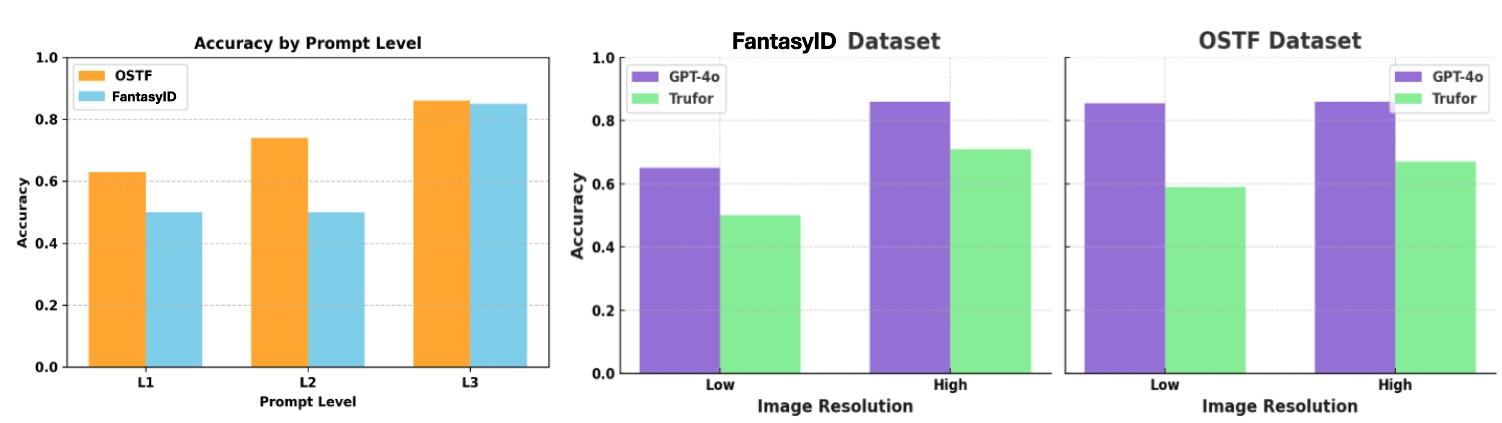}
    \caption{\textbf{Effect of Descriptive Prompts and Image Resolution:} (Left) With more detailed prompts, we can capture the task context well. (Right) Both GPT-4o and TruFor get a boost in their performance when image resolution is high. High resolution helps them capture smaller artifacts well.}
    \label{fig:promptvsacc}
\end{figure}
\vspace{-1.2em}
\paragraph{Prompts Matter.} The behavior of VLM is usually controlled by its prompts. This becomes crucial when we are evaluating its performance in zero-shot setting. We ablate the influence of prompts(temperature set as $0$) on GPT-$4$o responses . To this end, we design prompts with varying levels of description about the text manipulation task.
\begin{itemize}
    \item L1: this prompt only mentions that the query image may consist of some digital manipulations, without describing the nature of manipulation. The prompt is as follows:
   \begin{tcolorbox}[colback=gray!30, colframe=gray!30, boxrule=0pt, arc=4mm, left=1mm, right=1mm, top=1mm, bottom=1mm]
  You are an AI image manipulation detector. The image provided is taken in the wild and might have some digital manipulations.  
  Your task is to determine whether the image is manipulated or not?
  \end{tcolorbox}
   \item L2: Here, we mention that the text in the image can be manipulated and clearly mention the task is to determine whether the text regions are altered. The prompt is the following:
   \begin{tcolorbox}[colback=gray!30, colframe=gray!30, boxrule=0pt, arc=4mm, left=1mm, right=1mm, top=1mm, bottom=1mm]
  You are an AI image manipulation detector. The image provided is taken in the wild and might have text digitally altered.  
  Your task is to determine whether any of the text in the image is manipulated or not?
  \end{tcolorbox}

    \item L3: Lastly, we give a detailed description on the text manipulation task and add a few possible artifacts that can be attributed to the alteration process. This is exactly the prompt as in ~\cref{fig:ostf-prompt}.
\end{itemize}

\cref{fig:promptvsacc} summarizes the influence of prompts on GPT-$4$o prediction capability. The accuracy of GPT-$4$o model increases as the prompts get more detailed. For OSTF~\cite{qu2025revisiting} dataset, there is a consistent increase in the performance as prompts get detailed. However, for FantasyID, both L1 and L2 fail to capture the \emph{fantasy} nature of the ID cards and GPT-4o labels all the images as manipulated. This shows that performance can widely vary if prompts do not describe the underlying context of the task well.
\vspace{-1em}
\paragraph{Importance of Image Resolution.} Since text regions often occupy a relatively small area of the image, it is crucial that the model possesses sufficient representational power to capture such fine details. As discussed in~\cref{subsec:baselines}, the input image resolution varies across different models, which can significantly impact the performance on text manipulation detection tasks. Both FakeShield~\cite{xu2024fakeshield} and SIDA~\cite{huang2024sida} operate on lower-resolution image patches compared to Llama and Qwen models, limiting their ability to capture subtle artifacts. This limitation correlates with their lower performance, as shown in~\cref{tab:res_ostf}.
\begin{figure}[t]
    \centering
    \includegraphics[width=\linewidth]{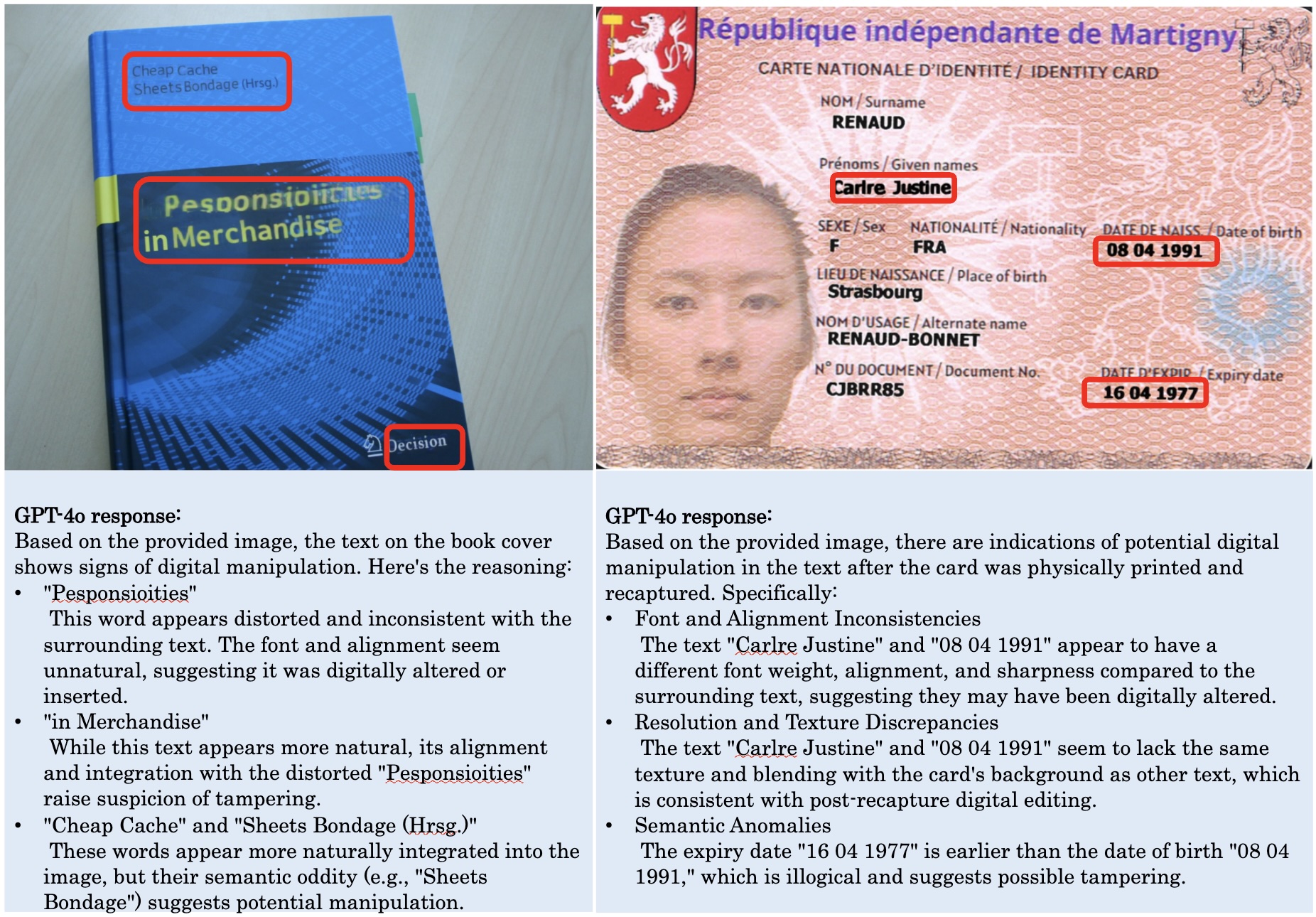}
    \caption{\textbf{GPT-4o response on OSTF and FantasyID:} We prompt GPT-4o to give a detailed response for its reasoning on text region manipulation. It is able to capture the manipulated regions(shown in \colorbox{red!60}{red}) based on inconsistency with the neighboring regions and other texts in the image. (Left) However, it does \emph{fail} to retrieve all the manipulated texts, bottom right text "Decision" is missing in its reasoning. (Right) Beyond texture and other pixel-wise artifacts, it can also detect semantic errors, as mentioned the expiry date is before birthdate.}
    \label{fig:gpt-ostf-res}
\end{figure}
To further illustrate this observation, we analyze the performance of the GPT-4o and TruFor~\cite{guillaro2023trufor} models by explicitly varying the input image resolution (see~\cref{subsec:baselines} for details). As shown in~\cref{fig:promptvsacc}, performance on both datasets improves when higher-resolution images are used. This effect is particularly significant in the case of the FantasyID dataset, where manipulation artifacts become indiscernible at lower resolutions, leading to random performance from TruFor~\cite{guillaro2023trufor}.


\vspace{-1em}
\paragraph{GPT-4o explanations.} GPT-$4$o performs much better compared to other the models. This is not surprising since it is the best performing VLM~\cite{chiang2024chatbot} on diverse benchmarks. To better understand reason for its performance, we prompt GPT-$4$o to follow~\cref{fig:ostf-prompt} and ask it to answer in detail. The corresponding response is shown in ~\cref{fig:gpt-ostf-res}. For both datasets, it generates an accurate description of textual tampering by analyzing variation in texture patterns, text alignment and semantic coherence in the image. Although, it identifies manipulated images well, it can still miss some manipulated regions. This issue can be better handled in the prompt by asking GPT-4o to single out all the manipulated words firsts and then individual verify for manipulation artifacts.

\section{Conclusion}
In this paper, we highlight the motivation and challenges of text manipulation detection. With our experiments, we bridge the gap of lack of text manipulation evaluations in the current literature. As VLMs are getting popular for image manipulation detection, we evaluate several models and can conclude the following: (a) The gap between closed source, GPT-4o and open source VLMs, Qwen-VL, Llama-Vision is large. The best performing open source model Qwen-VL trails behind GPT-4o by $7$\% on OSTF~\cite{qu2025revisiting} and $29$\% on FantasyID. To reduce this gap, the responses by GPT-4o  can be used to distill the knowledge into open source models and improve them.
(b) Specialized VLMs FakeShield~\cite{xu2024fakeshield} and SIDA~\cite{huang2024sida} fail to generalize to text manipulation. This highlights the challenges of text manipulation and the need of diverse training/evaluation datasets.
(c) VLM as a text manipulation detector works well when prompted with the right context and when they can process images in higher resolution. The underlying vision encoder and LLM affect the final performance.(d) FantasyID evaluation show that document manipulation detection is a challenging task, and most VLMs underperform.

We hope that research community will benefit from our findings and the proposed new dataset to improve VLMs for manipulation detection tasks.

\section{Acknowledgment}
This work was funded by InnoSuisse 106.729 IP-ICT ROSALIND project.

{\small
\bibliography{egbib}
}
\end{document}